\documentclass[10pt,journal,compsoc]{IEEEtran}

\usepackage{latexsym}
\usepackage{url}
\usepackage{booktabs}
\usepackage{times}
\usepackage{mathtools}
\usepackage{amsfonts}
\usepackage{graphicx}
\usepackage{nicefrac}
\usepackage{tabularx}
\usepackage{array}
\usepackage{color}
\usepackage{anyfontsize}
\usepackage{amsmath,enumitem}

\newcolumntype{Y}{X}

\ifCLASSOPTIONcompsoc
  \usepackage[nocompress]{cite}
\else
  \usepackage{cite}
\fi

\ifCLASSINFOpdf

\else

\fi

\begin{document}

\title{Sent2Matrix: Folding Character Sequences in Serpentine Manifolds for Two-Dimensional Sentence Representations}

\author{Hongyang Gao,
        Yi Liu,
        Xuan Zhang,
        and~Shuiwang~Ji,~\IEEEmembership{Senior Member,~IEEE}
        \IEEEcompsocitemizethanks{
          \IEEEcompsocthanksitem Hongyang Gao is with the
          Department of Computer Science, Iowa State University,
          Ames, IA 50011.\protect\\
          E-mail: hygao@iastate.edu 
          \IEEEcompsocthanksitem Yi Liu, Xuan Zhang, and Shuiwang Ji are with the
          Department of Computer Science and Engineering, Texas
          A\&M University, College Station, TX 77843.\protect\\
          E-mail: \{yiliu, xuan.zhang, sji\}@tamu.edu
          }
}

\markboth{}%
{Gao \MakeLowercase{\textit{et al.}}: Sent2Matrix: Folding Character
Sequences in Serpentine Manifolds for Two-Dimensional Sentence
Representations}

\IEEEtitleabstractindextext{%
\begin{abstract}
  We study text representation methods using deep models. Current
  methods, such as word-level embedding and character-level embedding
  schemes, treat texts as either a sequence of atomic words or a
  sequence of characters. These methods either ignore word
  morphologies or word boundaries. To overcome these limitations, we
  propose to convert texts into 2-D representations and develop the
  Sent2Matrix method. Our method allows for the explicit incorporation
  of both word morphologies and boundaries. When coupled with a novel
  serpentine padding method, our Sent2Matrix method leads to an
  interesting visualization in which 1-D character sequences are
  folded into 2-D serpentine manifolds. Notably, our method is the
  first attempt to represent texts in 2-D formats. Experimental
  results on text classification tasks shown that our method
  consistently outperforms prior embedding methods.
\end{abstract}

}

\maketitle

\IEEEdisplaynontitleabstractindextext

\IEEEpeerreviewmaketitle

\IEEEraisesectionheading{\section{Introduction}\label{sec:introduction}}

\IEEEPARstart{C}{onvolutional} neural
networks~(CNNs)~\cite{lecun1998gradient,simonyan2015very} have
achieved remarkable performance in various computer vision
tasks~\cite{imagenet_cvpr09,chen2016deeplab}. In natural language
processing~(NLP), recurrent neural
networks~(RNNs)~\cite{elman1990finding,hochreiter1997long,chung2014empirical}
were considered to be more natural, given that text data are
sequential. Recently, multiple NLP studies have shown that CNNs
can achieve competitive or even better performance than
RNNs~\cite{johnson2015effective,santos2014learning,johnson2015semi,gu2017recent}.
Other additional advantages of CNNs, including parallel and ease
of training, make CNNs even more attractive. Since CNNs take
structured data as input like images, raw text data need to be
converted into structured formats before applying CNNs.

Two popular text embedding methods are word-level
embedding~\cite{tang2014learning} and character-level
embedding~\cite{zhang2015character}. Word-level embedding methods
convert a word into a vector using a word dictionary and an
embedding matrix. Although they have been successfully applied to
various tasks~\cite{bahdanau2014neural,kim2014convolutional},
word-level embedding methods suffer from the limitation of
ignoring character-level morphologies~\cite{rekabsaz2017word}. In
character-level embedding method, each character is encoded into
an one-hot vector, thereby capturing relationships among
characters~\cite{kim2016character,vaswani2017attention}. On the
other hand, the importance of the word separator is weaken by
encoding it as a regular character in character-level embedding
methods.

In this work, we propose a new text representation method known
as the Sent2Matrix. Our method converts each word into a separate
matrix. Thus, a sentence can be represented as a 3-D tensor by
stacking the representations of words together, By explicitly
encoding word boundaries, our method enables CNNs to easily
capture both word-word and character-character relationships
simultaneously. To copy with variable-length words, we develop a
novel padding method known as the serpentine padding. Altogether,
our Sent2Matrix representation and serpentine padding methods
lead to an interesting visualization in which 1-D character
sequences are folded into 2-D serpentine manifolds using the word
separator as signal of direction change. Results on text
classification tasks demonstrate the effectiveness of our methods
compared to prior ones.

\section{Word-Level Embedding and Character-Level Embedding}

Deep learning methods take structured data as inputs. The most
common structure is in the form of regular grid, such as images.
When applying deep learning methods on texts, we need to convert
raw texts into some structured formats. Since text data
instances, such as sentences and documents, usually have variable
lengths, we also need to unify their lengths by appropriate
padding and trimming. Text data instances consist of words, which
in turn consist of characters. Thus, text data are commonly
converted into structured formats using either word-level
embedding~\cite{kim2014convolutional} or character-level
embedding~\cite{zhang2015character}. In this section, we
introduce these two commonly used embedding methods and describe
their limitations. To overcome these limitations, we propose the
Sent2Matrix embedding method. Our proposed method inherits the
advantages of both word-level and character-level embedding
methods while overcoming their limitations. In the following, we
use a fixed-size word vocabulary $V^{w}$ with size $|V^{w}|$
representing the number of words in the vocabulary.

\subsection{Word-Level Embedding}

The word-level embedding method considers each word as an atomic
entity and represents it as a single fixed-length vector.
Consider a sentence $s$ consisting of a sequence of $n$ words
\{$x_1, x_2, \cdots, x_n$\}, each word $x_i$, with $1 \le i \le
n$, is represented as an one-hot vector $\boldsymbol{u}^{i} \in
\mathbb{R}^{|V^{w}|}$ defined as
\begin{equation}\label{eq:word_one_hot}
{u}^{i}_j =
\begin{cases}
    1, & \text{if $V^{w}_j$ = $x_i$} \\
    0, & \text{otherwise},
\end{cases}
\end{equation}
where $u^i_j$ denotes the $j$-{th} element in $\boldsymbol{u}^i$.
Figure~\ref{fig:embedding} (a) provides an example of word-level
embedding. In many languages, the size of vocabulary $V^{w}$ is
very large, resulting in high-dimensional and sparse vectors
$\boldsymbol{u}^{i}$. These one-hot vectors cannot be used
directly and a dimensionality reduction step is performed to
project them into a lower-dimensional space as
$\boldsymbol{p}^{i} = {\boldsymbol M}^{w} \boldsymbol{u}^{i}$,
where $\boldsymbol M^{w}\in \mathbb{R}^{d^{w}\times |V^{w}| }$
denotes the word embedding matrix,
$\boldsymbol{p}^{i}\in\mathbb{R}^{d^{w}}$ denotes the reduced
representation, and $d^{w}$ is the reduced dimensionality.
Here, $M^{w}$ is a parameter matrix to be
learned from data, and $d^{w}$ is a user-specified parameter that
should depend on $|V^{w}|$ and complexity of tasks.

With these low-dimensional representations, a sentence $s$ with
length $n$ can be represented as a matrix as
\begin{equation}\label{eq:sen_word}
\boldsymbol S = [\boldsymbol{p}^{1}, \boldsymbol{p}^{2}, \cdots ,
\boldsymbol{p}^{n}]\in\mathbb{R}^{d^w\times n}.
\end{equation}
When the lengths of sentences are not equal to $n$, they need to
be zero-padded or trimmed appropriately. With such a fixed-length
vector sequence representation of sentences, the 1-D convolution
operation is usually applied to compute high-level features. In
this operation, the embedding dimension is treated as the channel
dimension, and the 1-D convolution is applied along the word
dimension. Specifically, the 1-D convolution operation uses a set
of 1-D learnable filters $\{\boldsymbol{w}_j \in
\mathbb{R}^{h}\}_{j=1}^{d^{w}}$ to compute output features, where
$k$ denotes the 1-D kernel size. For example, a feature value
$\boldsymbol{y}_i \in \mathbb{R}$ is computed as
\begin{equation}\label{eq:word_conv}
\boldsymbol y_i = f\left(\sum_{j=1}^{d^w}\boldsymbol{w}_j \odot \boldsymbol S^{j}_{i:i+h-1} +
\boldsymbol b \right),
\end{equation}
where $\odot$ denotes element-wise multiplication, $\boldsymbol
S^{j}_{i:i+h-1}$ denotes the column vector including elements in
the $j$th row and the $i$th to the $(i+h-1)$-th columns of
$\boldsymbol S$ in order, $\boldsymbol b$ denotes the bias, and
$f(\cdot)$ is a non-linear function such as the
ReLU~\cite{nair2010rectified}. Suppose the stride in convolution
is set to $1$, the same set of filters are applied to every
possible word window of size $h$ in the sentence, resulting in
the following output feature vector:
\begin{equation}
\boldsymbol{y} = [y_1, y_2, \cdots, y_{n-h+1}]^T\in\mathbb{R}^{n-h+1}.
\end{equation}\label{eq:word_conv_re}
We can use multiple sets of independent filters to compute
multiple output feature vectors. These feature vectors form the
different channels of features to be used as inputs for the next
layer.

Although the word-level embedding method has achieved great
success in various tasks such as neural machine
translation~\cite{bahdanau2014neural,luong2015effective} and text
classification~\cite{tang2014learning}, it suffers from several
limitations. In particular, this method considers words as atomic
representations and does not explicitly incorporates their
character constitutions and morphologies, such as roots,
prefixes, and suffixes, in learning the representations. Each
word is discretized to a one-hot representation, and the semantic
relations among words are inferred only from contexts. For
example, the words ``surprise'' and ``surprising'' have similar
meanings, but their one-hot representations are not related. The
similarity of their low-dimensional embedding needs to be
inferred based on the contexts in which they are used.

\subsection{Character-Level Embedding}

\begin{figure*}[t]
  \includegraphics[width=\textwidth]{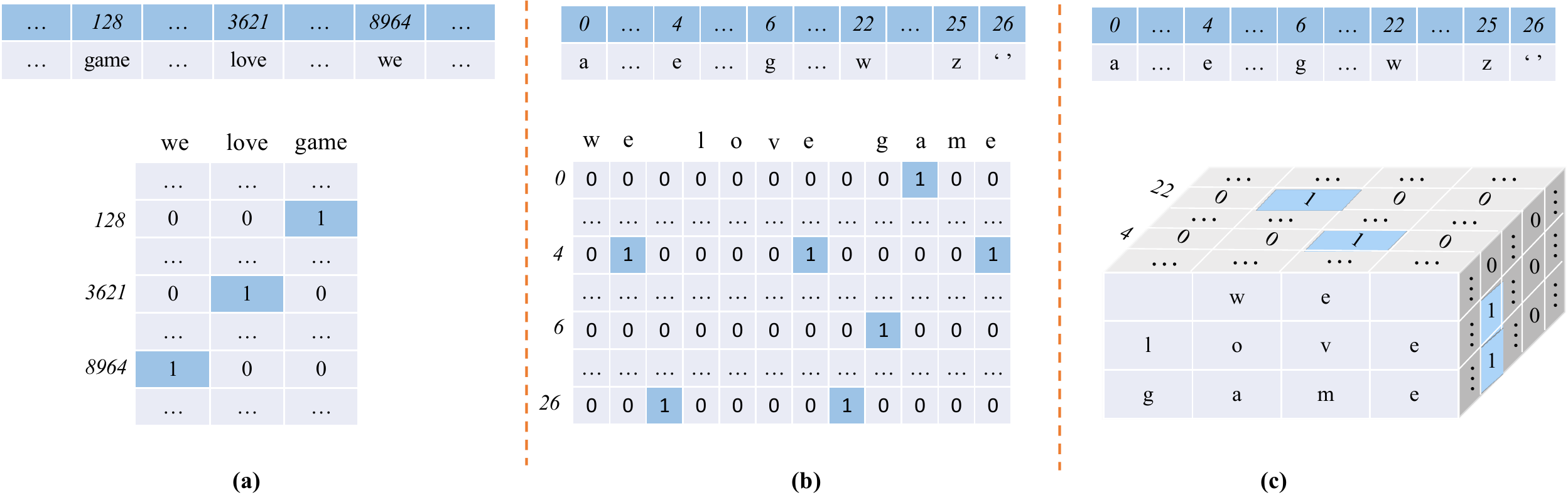}
  \caption{Comparison of three embedding methods. Figures (a),
  (b), and (c) show the representations of the same sentence
  using word-level, character-level, and Sent2Matrix embedding
  methods, respectively.} \label{fig:embedding}
\end{figure*}

To overcome the limitations of word-level embedding, the
character-level embedding method was proposed to consider
morphological information explicitly~\cite{zhang2015character}.
In this method, a sentence is considered as a sequence of
characters, and each character is encoded into an one-hot vector
based on a character-level vocabulary $V^{c}$. Given a sentence
$s$ consisting of a sequence of $m$ characters $\{z_1, z_2,
\cdots, z_m \}$, the one-hot vector
$\boldsymbol{v}^i\in\mathbb{R}^{|V^{c}|}$ for character $z_i$ can
be expressed as
\begin{equation}\label{eq:char_one_hot}
\boldsymbol{v}^{i}_{j} =
\begin{cases}
    1, & \text{if $V^{c}_j$ = $z_i$} \\
    0, & \text{otherwise}.
\end{cases}
\end{equation}
It is worth noting that the size of the character-level
vocabulary is usually much smaller than that of the word-level
vocabulary; that is, $|V^{c}| \ll |V^{w}|$. Hence, the one-hot
vectors in character-level embedding can be used directly without
employing the embedding matrix as in the word-level embedding
method. The sentence $s$ consisting of $m$ characters can be
represented as a matrix as
\begin{equation}\label{eq:sen_char}
\boldsymbol T = [\boldsymbol{v}^{1}, \boldsymbol{v}^{2},\cdots,
\boldsymbol{v}^{m}]\in\mathbb{R}^{|V^c|\times m}.
\end{equation}
Figure~\ref{fig:embedding} (b) provides an example of
character-level embedding. Similar to the case of word-level
embedding, 1-D convolutions are applied on $T$ to compute high-level
features.

It can be seen from the descriptions above that the convolution
operation considers the relationships among characters, thereby
extracting features with explicit morphological information. In
character-level embedding, a sentence is considered as a sequence
of characters. The separator between words is commonly a specific
character such as a space. However, such kind of separators make
the boundaries between words vague from the view of subsequent
neural network operations. For CNNs, there is no difference
between the embedding vectors of separators and those of other
characters. Thus, word boundaries are not explicitly given to
subsequent neural network operations. Hence, CNNs need to extract
more advanced features to consider the relationships between
words.

\section{Sent2Matrix}

In order to overcome the limitations of word-level and
character-level embedding methods, we propose the Sent2Matrix
embedding method. Our proposed Sent2Matrix method can consider
the relationships among characters and those among words
simultaneously. When combined with a novel padding method
described below, our methods lead to a new way of converting
character streams into a 2-D representation on which 2-D neural
network operations can be applied.

\subsection{Sent2Matrix Representations}

Given a sentence $s=\{x_1,x_2,\cdots,x_n\}$ consisting of at most
$n$ words in which each word $x_i$ contains at most $m$
characters, we first encode each character into a one-hot vector
using a character-level vocabulary. Unlike the vocabulary used in
the regular character-level embedding above, we do not need to
include the word separator in vocabulary. For notational
convenience, we denote this character-level vocabulary as $V^c$
again. Each character $z_{i, j}$ for ($1 \le i \le n,\, 1 \le j
\le m$) is encoded into a one-hot vector $\boldsymbol{v}^{i,
j}\in\mathbb{R}^{|V^c|}$ in a way that is similar to the case of
regular character-level embedding as
\begin{equation}\label{eq:sent_one_hot} \boldsymbol{v}^{i, j}_k = \begin{cases}
1, & \text{if $V^{c}_k$ = $z_{i, j}$} \\     0, & \text{otherwise}.
\end{cases} \end{equation}

With the character-level encoding described as above, our
proposed Sent2Matrix embedding encodes each word as a separate
matrix. Then the sequence of words in a sentence can be
represented as a 3-D data array, known as a 3-D
tensor~\cite{Tensor:Tamara}. By using this higher-dimensional
representation of sentences, the word boundaries have been
encoded explicitly while the word morphologies have also been
considered in the character-level encoding. In particular, each
word $x_i$ can be represented as a matrix as
\begin{equation}\label{eq:sent_word}
  \boldsymbol X_i = [\boldsymbol{v}^{i, 1}, \boldsymbol{v}^{i, 2}, \dots \boldsymbol{v}^{i,
m}]\in\mathbb{R}^{|V^c|\times m}.
\end{equation}
Since $m$ is the maximum number of characters in a word, appropriate
padding strategies are needed to deal with variable-length words. In
fact, novel padding strategies and related interpretations are the
other major contributions of this work. These details will be given
below.

\begin{figure*}[t]
  \includegraphics[width=\textwidth]{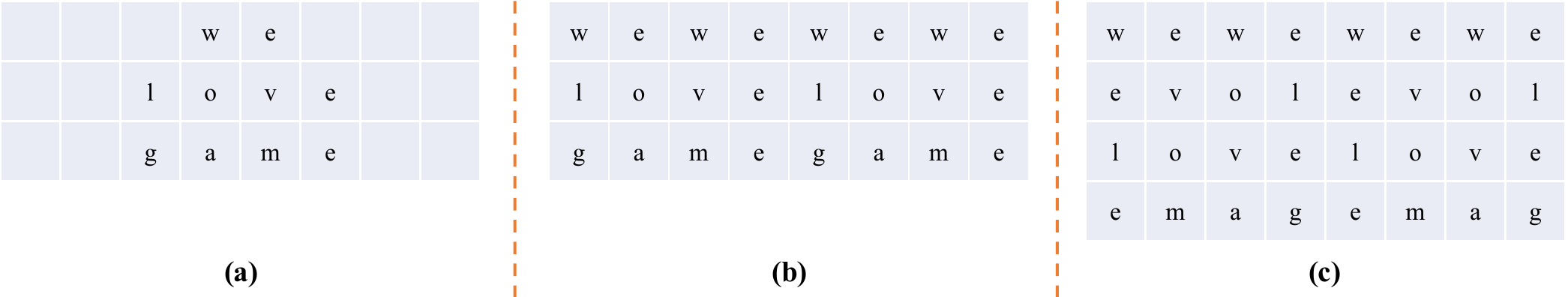}
  \caption{Illustrations of the three padding methods on the same
  sentence. Figures (a) and (b) show the encoded results using zero
  padding and cyclic padding, respectively. Figure (c) describes the
  encoded output with serpentine padding. We repeat each word twice
  except for the first and the last words. The first occurrence of
  each repeated words and the last word are in reverse order. Cyclic
  padding was used for words that are shorter than the required
  length.}\label{fig:padding}
\end{figure*}

Based on the above word representations, the sentence $s$ can be
represented as a 3-D tensor $\mathcal{S}$ as:
\begin{equation}\label{eq:sen_sent}
\boldsymbol{\mathcal{S}} = [\boldsymbol X_1, \boldsymbol X_2,\cdots, \boldsymbol X_n]\in\mathbb{R}^{n\times m\times
|V^c|}.
\end{equation}
In Eq.~(\ref{eq:sen_sent}), the matrices $\boldsymbol X_i$ are
stacked to form the tensor $\boldsymbol{\mathcal{S}}$ by treating
$\boldsymbol X_i$ as the $i$th horizontal slice of
$\boldsymbol{\mathcal{S}}$~\cite{Tensor:Tamara}. An example is
given in Figure~\ref{fig:embedding} (c) to illustrate the
Sent2Matrix embedding of a sentence. By this representation, a
sentence $s$ is now encoded into a 3-D tensor
$\boldsymbol{\mathcal{S}}$ in which the three dimensions
correspond to word, character, and character embedding,
respectively.

Recall that only 1-D convolutions have been applied to compute
high-level features in word-level and character-level embedding
methods. In contrast, our proposed Sent2Matrix embedding method
represents text data in an image-like format, thereby enabling
the use of 2-D convolutions to compute features that capture
relationships among both characters and words simultaneously.
Given the tensor representation of a sentence $\mathcal{S}$ we
can apply a 2-D convolution operation using a set of learnable
filters $\{\boldsymbol W_i \in \mathbb{R}^{k_1 \times
k_2}\}_{i=1}^{|V^c|}$, where $k_1$ and $k_2$ represent the sizes
of filter. Then an output feature value can be computed as
\begin{equation}\label{eq:sent_conv}
y_{i,j} = f\left(\sum_{\ell=1}^{|V^c|}
\boldsymbol W_{\ell}\odot \boldsymbol{\mathcal{S}}_{i:i+k_1-1, j:j+k_2-1,\ell} + b\right),
\end{equation}
where the subscripts for $\boldsymbol{\mathcal{S}}$ denote taking
the corresponding elements in it along each of the three
dimensions as in~\cite{Tensor:Tamara}.

\begin{figure}[t]
\includegraphics[width=\linewidth]{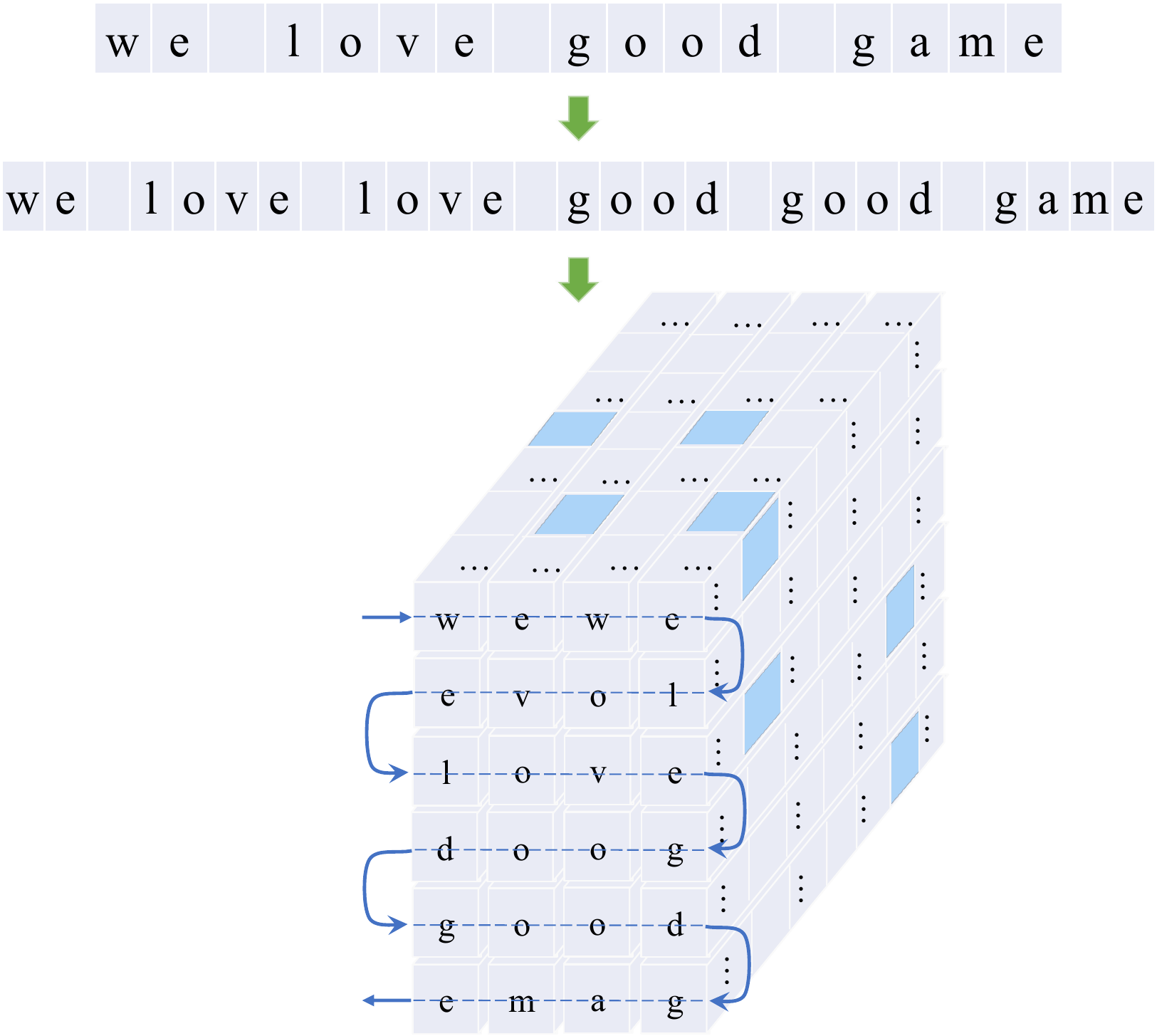} \caption{A
visualization of the serpentine padding. A sentence with repeated
words as described in section~\ref{sec:sent_pad} is folded from
left to right and then right to left in serpentine manifolds
using space as the U-turn signal.} \label{fig:serp}
\end{figure}

By computing features on each possible patch of size $k_1\times
k_2$ using the same set of filters $W_i$, we can obtain an output
feature matrix $\boldsymbol Y\in\mathbb{R}^{(m-k_1+1)\times
(n-k_2+1)}$ as
\begin{equation}\label{eq:sent_conv_re1}
  \boldsymbol Y =
\begin{bmatrix}
    y_{11}  & \dots & y_{1,n-k_2+1} \\
    y_{21} & \dots & y_{2,n-k_2+1} \\
    \vdots  & \ddots & \vdots \\
    y_{m-k_1+1,1} & \dots & y_{m-k_1+1,n-k_2+1}
\end{bmatrix}.
\end{equation}
We can use multiple sets of independent filters to compute multiple
output feature matrices. These feature matrices form the different
channels of features to be used as inputs for the next layer. By
using 2-D filters for feature extraction, Sent2Matrix can capture
the relationships among characters from different words, thereby
providing additional morphological information for feature
extraction.

\subsection{Sent2Matrix Padding}\label{sec:sent_pad}

\begin{figure*}[t] \includegraphics[width=\linewidth]{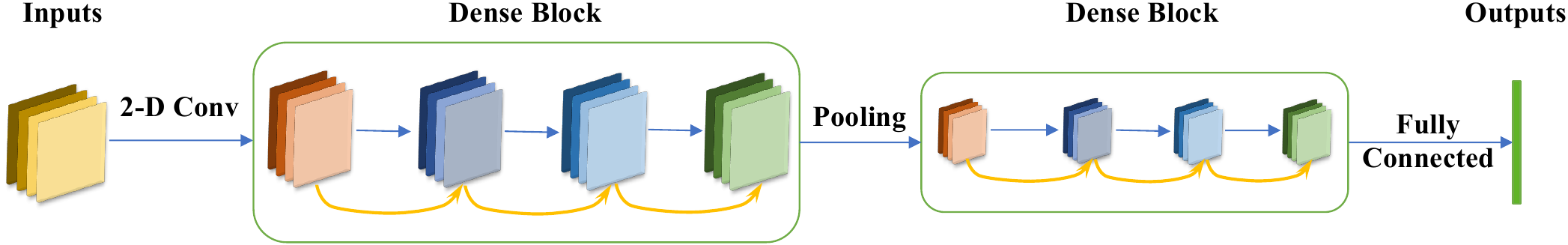}
\caption{The architecture of a sample dense network for text
classification. This network includes two dense blocks and employs a
fully-connected layer for final prediction.} \label{fig:dense}
\end{figure*}

It follows from the above description of Sent2Matrix representation that a
sentence has at most $n$ words and each word is assumed to have the same
length $m$. For sentence with less than $n$ words, we center the words and pad
zero matrices at two ends. At word level, we use $m$ as the maximum length of
words in a sentence and propose to apply advanced padding strategies to
convert shorter words to contain exactly $m$ characters. In this section, we
develop three strategies for padding.

\textbf{Zero Padding:} Padding is required in convolution layers of
CNNs to keep the size of feature maps, and zero-padding is the most
commonly used form of padding. In this strategy, we consider the
property of convolution operation and propose to center the
characters of words in the middle of the vector and pad zeros at two
ends. This is because elements in the middle will be covered by more
convolution windows when performing convolution, thereby leading to
effective use of the input data and yielding more informative features.
The zero padding strategy is illustrated in Figure~\ref{fig:padding}
(a). However, since lengths of sentences vary, this strategy may
introduce many zero values and waste spaces in resulting embeddings.

\textbf{Cyclic Padding:} Although the zero-padding strategy centers
characters for better convolution coverage, only a small number of
convolution windows can cover characters in very short words such as
``I''. In this case, the padded zeros become some kind of noises and
compromise CNNs' performance. To overcome this limitation of the
zero-padding strategy, we propose the cyclic padding strategy. In
this strategy, the characters are repeated a number of times until
the maximum length $m$ is reached. Compared to zero-padding, cyclic
padding employs the contents of words in the padded positions,
thereby allowing all filter windows to access the contents of words.
Figure~\ref{fig:padding} (b) provides an example of cyclic padding.

\textbf{Serpentine Padding:} The character-level embedding considers
a sentence as a sequence of characters. Specifically, 1-D
convolutional filters can capture the relationship between the
trailing characters of one word and the leading characters of the
following word. We observe that both the zero padding and cyclic
padding fail to preserve the sequential order of characters between
adjacent words. To restore the sequential flow of character stream
in sentence, we propose the serpentine padding strategy. In this
strategy, every word in the sentence is repeated twice except for
the first and the last words. The second occurrence of each word is
in normal order, while the first occurrence of each word is in
reverse character order. In addition, the last word only appears in
reverse character order.

Given a sentence $s=\{x_1,x_2,\cdots,x_n\}$ that consists of $n$
words, the repeated sentence $\tilde s$ can be written as $\tilde
s=\{x_1,\overleftarrow{x_2},x_2,\overleftarrow{x_3},x_3,\cdots,\overleftarrow{x_{n-1}},x_{n-1},\overleftarrow{x_n}\}$,
where $\overleftarrow{x_i}$ denotes the word $x_i$ in reverse
character order. Each individual word, including the repeated
ones, is padded to include $m$ characters using cyclic padding.
Hence, the repeated sentence $\tilde s$ can be represented as a
3-D tensor as $\tilde{\mathcal{S}}\in\mathbb{R}^{2(n-1)\times
m\times |V^c|}$ using our proposed Sent2Matrix representation. We
propose to use a stride of $2$ along the word dimension when
applying convolution operations to compute high-level features
from $\tilde{\mathcal{S}}$. This forces the filters to compute
features from adjacent words in the character order in the
original sentence. Figure~\ref{fig:padding} (c) provides an
illustration of using our serpentine padding with the proposed
Sent2Matrix representation of a text.

\subsection{Interpretation of Serpentine Padding as Folding of Character Sequences in 2-D}

In the previous section, we propose the serpentine padding
method. The proposed method above is motivated by an intention to
fold a sequence of characters representing a sentence into a
serpentine pattern using the word separator as the signal for
making a $180^{\circ}$ turn (Figure~\ref{fig:serp}). This view
can be best understood by considering a different version of the
repeated sentence $\bar
s=\{x_1,x_2,x_2,x_3,x_3,\cdots,x_{n-1},x_{n-1},x_n\}$ in which
each word is padded to have $m$ characters using the proposed
cyclic padding. When the 3-D tensor $\tilde{\mathcal{S}}$ is
considered as a 2-D array of mode-3 fibers
$\tilde{\mathcal{S}}_{ij:}$~\cite{Tensor:Tamara},
$\tilde{\mathcal{S}}$ can be obtained by filling in the fibers
using the character embedding vectors $\boldsymbol{v}^{i, j}$,
defined in Eq.~(\ref{eq:sent_one_hot}), for $\bar s$. In
particular, these fibers are filled in using the character
embedding vector sequence representation of a sentence from left
to right and then right to left in serpentine pattern using the
word separator as the signal for making a $180^{\circ}$ turn. 

\begin{table*}[t]
  \caption{Summary of statistics for the 4 datasets. The {$n$}
      and {$m$} values for each dataset are also given, where $n$
      and $m$ denote the maximum number of words in a sentence
      and the maximum number of characters in a word.}
  \label{table:datasets}
  \begin{tabularx}{\linewidth}{ l YYYY Y }
    \hline
    \textbf{Datasets}                   & \textbf{\#Classes} & \textbf{\#Training Samples} & \textbf{\#Testing Samples} & $\boldsymbol{n}$ & $\boldsymbol{m}$ \\ \hline
    \hline
    AG's News~\cite{zhang2015character} & 4                  & 120,000                     & 7,600                      & 49               & 18               \\ \hline
    Yelp Full~\cite{zhang2015character} & 5                  & 650,000                     & 50,000                     & 67               & 18               \\ \hline
    MR~\cite{pang2005seeing}            & 2                  & 10,235                      & 427                        & 51               & 18               \\ \hline
    \hline
  \end{tabularx}
\end{table*}

\subsection{Position Embedding}

Text data are sequential but the position information of entities
is not explicitly modeled in CNNs. This is because CNNs do not
consider sequential information explicitly. Thus, it is desirable
to explicitly encode position information. To capture position
information, an embedding method was proposed
in~\cite{zeng2014relation} to encode the positions of words or
characters in sentences. It has been shown to be effective for
various
tasks~\cite{dos2015classifying,feng2015applying,zhang2017position}.
The position embedding can help to encode relative position
information of characters in the text. To use position
information in our method, we encode the positions of characters
in words into one-hot vectors and concatenate them with
character-level encoding vectors.

\subsection{Network Design}

In our proposed Sent2Matrix embedding method, each text is
converted into the 2-D format, which enables the usage of 2-D
convolution operations on text data. The usage of 2-D convolution
operation can help to capture dependencies among characters of
different words. To this end, we build a densely-connected
network for text data.

Given an input text, we first convert it into a 3-D tensor. We
apply a 2-D convolution layer with stride 2 in the word dimension
to encode high-level features. After that, we stack several
densely-connected blocks. In each block, there are multiple
convolution layers. We use concatenation to combine the input and
output of each layer. An average pooling layer is used between
each pair of consecutive blocks. The output feature maps of the
final block are flattened and fed into a two-layer feed-forward
neural network for prediction. Figure~\ref{fig:dense} provides a
simple example of our densely-connected network.

\section{Related Work}

Many studies have applied CNNs on text classification tasks using
word-level or character-level embedding methods. A simple CNN model
based on word-level embedding was proposed in
\cite{tang2014learning} to improve the performance on sentiment
analysis and question classification tasks. In
\cite{zhang2015character}, the authors employed character-level
embedding method to build convolutional networks for text
classification. Both word-level and character-level embedding
methods were extensively used in other NLP tasks such as neural
machine translation~\cite{bahdanau2014neural,chung2016character}.

In addition to these two embedding methods, some studies tried
other ways for text transformation. In~\cite{sennrich2016neural},
the authors proposed subword units to address rare words in
open-vocabulary problems. Subword units can be considered as a
trade-off between word-level and character-level embedding
methods. \cite{dos2014deep} employed a neural network to encode
characters of each word into a character-level embedding vector.
This vector was concatenated with word-level embedding to form
the final word representation. This method can be seen as an
attempt of combining word-level and character-level embedding.
However, the character-level embedding in this work only focus on
the characters within each word. Our method enables CNNs to
consider relationships among characters across words.

\cite{howard2018universal} proposed a pretraining method to
train a network for text classification.

\textbf{employed a neural network to encode characters of each
word into a character-level embedding vector. This vector was
concatenated with word-level embedding to form the final word
representation. This method can be seen as an attempt of
combining word-level and character-level embedding. However, the
character-level embedding in this work only focus on the
characters within each word. Our method enables CNNs to consider
relationships among characters across words
Our method enables CNNs to consider
relationships among characters across words
relationships among characters across words
Our method enables CNNs to consider
relationships among characters across words
Our method enables CNNs to consider}

\section{Experimental Studies}

We evaluate our proposed Sent2Matrix method on text classification
tasks using CNNs as a basic model architecture. We conduct
experiments to compare with several baseline methods, including
word-level CNNs, character-level CNNs and long short-term memory
(LSTM) RNNs. Some results of these baseline methods are reported
in~\cite{kim2014convolutional} and~\cite{zhang2015character}. In
addition, performance studies are used to compare the three padding
strategies. Our results show that the proposed Sent2Matrix
outperforms these prior state-of-the-art methods, and the serpentine
padding yields improved performance.

\begin{table*}[t]
  \centering \caption{Results of text classification experiments
    in terms of classification accuracy on the AG's News, Yelp
    Full, and MR datasets. ``Sent2Matrix CNN'' denotes the CNN
    based on our Sent2Matrix embedding method. $^\star$ indicates
    results obtained by ourselves using the same settings as the
    Sent2Matrix CNN. For LSTM results obtained by ourselves, we
    employ the same architecture as in~\cite{zhang2015character}.
    The result labeled by “-” is not available.}
  \label{table:models}
  \begin{tabularx}{\linewidth}{ lYYY }
    \hline
    \textbf{Model}                                          & \textbf{AG's News} & \textbf{Yelp Full} & \textbf{MR}     \\ \hline
    \hline
    LSTM~\cite{zhang2015character}                          & 86.1\%             & 58.2\%             & 74.6\%$^\star$  \\ \hline
    Word-Level CNN w/o Word2Vec~\cite{kim2014convolutional} & 85.9\%$^\star$     & 57.1\%$^\star$     & 76.1\%          \\ \hline
    Word-Level CNN w/ Word2Vec~\cite{zhang2015character}    & 91.4\%             & 60.4\%             & -               \\ \hline
    Char-Level CNN~\cite{zhang2015character}                & 87.2\%             & 62.1\%             & 76.8$^\star$    \\ \hline
    \textbf{Sent2Matrix CNN}                                & \textbf{92.1\%}    & \textbf{63.2\%}    & \textbf{77.9\%} \\ \hline
    \hline
  \end{tabularx}
\end{table*}

\subsection{Datasets}

We evaluate our methods on three datasets involving two types of
text classification tasks; namely topic classification and
sentiment classification. We choose two large datasets and one
small dataset in terms of sample size. The statistics for these
datasets are summarized in the first 3 columns of
Table~\ref{table:datasets}. For datasets that training and test
split was not given, we randomly split them into training and
test sets to ensure all methods use the same training data for
fair comparisons.

\textbf{AG's News} is a topic classification
dataset~\cite{zhang2015character} containing four topics: World,
Sports, Business and Sci/Tech. AG is a collection containing more
than 1 million news articles, and the final dataset is formed by
choosing four classes, each containing 30,000 training samples
and 1,700 test samples. Each sample is a short text consisting of
several sentences. The label indicates the sentiment of a short
text.

\textbf{Yelp Full} is obtained from Yelp Dataset Challenge in 2015
and compiled by~\cite{zhang2015character}. The dataset is for
sentiment classification, and it includes five classes indicating
movie review star from 1 to 5. Each class contains 130,000 training
samples and 10,000 test samples. Each sample is a short text.

\textbf{MR} is a Movie Review dataset~\cite{pang2005seeing}, and the
task is for sentiment classification containing positive and
negative reviews. Each sample is a short sentence, and the longest
sentence contains 51 words.

\subsection{Experimental Setup}

We use two sets of experimental settings for the three datasets due
to their different sizes. On the small datasets MR,
we use the same model architectures as described
in~\cite{kim2014convolutional} with minor changes to accommodate the
Sent2Matrix embedding. On the large datasets AG's News and Yelp
Full, we build a new network based on densely connected
convolutional networks~(DCNNs)~\cite{huang2017densely}. In the
following, we mainly discuss the experimental settings on large
datasets. These for MR are provided
in~\cite{kim2014convolutional}.

\textbf{Choice of vocabulary:} In our proposed Sent2Matrix
embedding method, each character is encoded by a pre-determined
character vocabulary. The size of vocabulary is an important
hyper-parameter, which provides a trade-off between
representational capacity and computational efficiency. Common
elements of character vocabulary include lower-case characters,
upper-case characters, and punctuations. We observe that
case-sensitivity and punctuations do not contribute much to
prediction. For instance, ``GOOD'' and ``good'' should lead to
the same prediction in sentiment classification task using the
Yelp Full dataset. From this point, our character vocabulary only
contains 26 lower-case characters, which also facilitates the
training process.

\textbf{Choices of padding parameters:} In addition to the character
vocabulary, we have another two hyper-parameters; namely the maximum
number of words in sentence $n$ and maximum length of words $m$.
Given the statistics of texts in the two large datasets, we set $m$
to 18, which covers 99\% of the words in datasets. For maximum words
number $n$, we use 49 and 67 for AG's News and Yelp Full,
respectively. The values of $m$ and $n$ for each dataset are given
in Table~\ref{table:datasets}.

For both the small and large datasets, the following settings are
shared. For all layers, we use ReLU~\cite{nair2010rectified} as the
activation function with a dropout keep rate of 0.5. For training,
we use the Adam optimizer~\cite{kingma2014adam} with a learning rate
of 0.001. The mini batch size used for all datasets is 512. All
hyper-parameters are tuned based on the validation datasets of MR
and AG's News.

\subsection{Comparison of Padding Strategies}

\begin{table}[t]
  \caption{Comparison among the three padding methods on the AG's
    News and Yelp Full datasets.} \label{table:paddings}
  \begin{tabularx}{\linewidth}{  l  Y Y }
    \hline
    \textbf{Padding Method}     & \textbf{AG's News} & \textbf{Yelp Full} \\ \hline
    \hline
    Zero Padding                & 88.5\%             & 58.7\%             \\ \hline
    Cyclic Padding              & 90.1\%             & 60.2\%             \\ \hline
    \textbf{Serpentine Padding} & \textbf{92.1\%}    & \textbf{63.2\%}    \\ \hline
    \hline
  \end{tabularx}
\end{table}

We compare the performance of the three proposed padding
strategies on the AG's News dataset, and the results are
summarized in Table~\ref{table:paddings}. We can observe from the
results that the cyclic padding strategy outperforms the zero
padding by 1.6\%, which confirms the effectiveness of making
words content available for all filter windows. The serpentine
padding outperforms the zero padding and the cyclic padding by a
margin of 2.0\% and 2.6\% on AG's News dataset, respectively.
This demonstrates the benefits of preserving the sequential order
of character stream by using the proposed serpentine padding
strategy. The following experiments will only use the serpentine
padding method.

\subsection{Comparison of Sent2Matrix with Other Methods}

\begin{table*}[t]
  \centering \caption{Results of text classification experiments
    in terms of classification accuracy on the AG's News, Yelp
    Full, and MR datasets.}
  \label{table:pretrain}
  \begin{tabularx}{\linewidth}{ lYYY }
    \hline
    \textbf{Model}                                          & \textbf{AG's News} & \textbf{Yelp Full} & \textbf{MR}     \\ \hline
    \hline
    LSTM~\cite{zhang2015character}                          & 86.1\%             & 58.2\%             & 74.6\%$^\star$  \\ \hline
    Word-Level CNN w/o Word2Vec~\cite{kim2014convolutional} & 85.9\%$^\star$     & 57.1\%$^\star$     & 76.1\%          \\ \hline
    Word-Level CNN w/ Word2Vec~\cite{zhang2015character}    & 91.4\%             & 60.4\%             & -               \\ \hline
    Char-Level CNN~\cite{zhang2015character}                & 87.2\%             & 62.1\%             & 76.8$^\star$    \\ \hline
    \textbf{Sent2Matrix CNN}                                & \textbf{92.1\%}    & \textbf{63.2\%}    & \textbf{77.9\%} \\ \hline
    \hline
  \end{tabularx}
\end{table*}

\begin{table*}[t]
  \centering \caption{Results of text classification experiments
    in terms of classification accuracy on the AG's News, Yelp
    Full, and MR datasets.}
  \label{table:attn}
  \begin{tabularx}{\linewidth}{ lYYY }
    \hline
    \textbf{Model}                                          & \textbf{AG's News} & \textbf{Yelp Full} & \textbf{MR}     \\ \hline
    \hline
    LSTM~\cite{zhang2015character}                          & 86.1\%             & 58.2\%             & 74.6\%$^\star$  \\ \hline
    Word-Level CNN w/o Word2Vec~\cite{kim2014convolutional} & 85.9\%$^\star$     & 57.1\%$^\star$     & 76.1\%          \\ \hline
    Word-Level CNN w/ Word2Vec~\cite{zhang2015character}    & 91.4\%             & 60.4\%             & -               \\ \hline
    Char-Level CNN~\cite{zhang2015character}                & 87.2\%             & 62.1\%             & 76.8$^\star$    \\ \hline
    \textbf{Sent2Matrix CNN}                                & \textbf{92.1\%}    & \textbf{63.2\%}    & \textbf{77.9\%} \\ \hline
    \hline
  \end{tabularx}
\end{table*}

We compare Sent2Matrix CNN with other state-of-the-art models. The
experimental results are summarized in Table~\ref{table:models}. We
can see that Sent2Matrix CNN outperforms word-level CNN and
character-level CNN by at least a margin of 0.7\%, 1.1\%, and 1.1\%
on the AG's News, Yelp Full, and MR datasets,
respectively. Also, the margins tend to be larger for larger
datasets. These results provide some insights about our embedding
method. On one hand, the promising performance of our model on the
small datasets demonstrate the representational ability our method
compared to word-level and character-level embedding methods. On the
other hand, the advantages of Sent2Matrix on large datasets are even
more remarkable than that on small datasets. This indicates that the
Sent2Matrix embedding method enables CNN to apply 2-D filters to
compute more advanced features, thereby leading to better
generalization. In addition, all CNN models, including Sent2Matrix
CNN, achieve better performance than that of LSTM on all datasets.
This is consistent with recent results in other studies and
demonstrates the effectiveness of CNNs compared to RNNs. These
results show that our proposed method yields consistently better
performance across all datasets. This clearly demonstrates the
effectiveness of modeling texts using two-dimensional matrices in
the proposed Sent2Matrix embedding.

\subsection{{Performance Study using Pretraining}}

\textbf{Network design.}
We compare Sent2Matrix CNN with other state-of-the-art models. The
experimental results are summarized in Table~\ref{table:models}. We
can see that Sent2Matrix CNN outperforms word-level CNN and
character-level CNN by at least a margin of 0.7\%, 1.1\%, and 1.1\%
on the AG's News, Yelp Full, and MR datasets,
respectively. Also, the margins tend to be larger for larger
datasets. These results provide some insights about our embedding
method. On one hand, the promising performance of our model on the
small datasets demonstrate the representational ability our method
compared to word-level and character-level embedding methods. On the
other hand, the advantages of Sent2Matrix on large datasets are even
more remarkable than that on small datasets. This indicates that the
Sent2Matrix embedding method enables CNN to apply 2-D filters to

\textbf{Results.}
We compare Sent2Matrix CNN with other state-of-the-art models. The
experimental results are summarized in Table~\ref{table:models}. We
can see that Sent2Matrix CNN outperforms word-level CNN and
character-level CNN by at least a margin of 0.7\%, 1.1\%, and 1.1\%
on the AG's News, Yelp Full, and MR datasets,
respectively. Also, the margins tend to be larger for larger
datasets. These results provide some insights about our embedding
method. On one hand, the promising performance of our model on the
small datasets demonstrate the representational ability our method
compared to word-level and character-level embedding methods. On the
other hand, the advantages of Sent2Matrix on large datasets are even
more remarkable than that on small datasets. This indicates that the
Sent2Matrix embedding method enables CNN to apply 2-D filters to
compute more advanced features, thereby leading to better
generalization.

\subsection{{Performance Study using Attention Operator}}

\textbf{Network design.}
We compare Sent2Matrix CNN with other state-of-the-art models. The
experimental results are summarized in Table~\ref{table:models}. We
can see that Sent2Matrix CNN outperforms word-level CNN and
character-level CNN by at least a margin of 0.7\%, 1.1\%, and 1.1\%
on the AG's News, Yelp Full, and MR datasets,
respectively. Also, the margins tend to be larger for larger
datasets. These results provide some insights about our embedding
method. On one hand, the promising performance of our model on the
small datasets demonstrate the representational ability our method
compared to word-level and character-level embedding methods. On the
other hand, the advantages of Sent2Matrix on large datasets are even
more remarkable than that on small datasets. This indicates that the
Sent2Matrix embedding method enables CNN to apply 2-D filters to

\textbf{Results.}
We compare Sent2Matrix CNN with other state-of-the-art models. The
experimental results are summarized in Table~\ref{table:models}. We
can see that Sent2Matrix CNN outperforms word-level CNN and
character-level CNN by at least a margin of 0.7\%, 1.1\%, and 1.1\%
on the AG's News, Yelp Full, and MR datasets,
respectively. Also, the margins tend to be larger for larger
datasets. These results provide some insights about our embedding
method. On one hand, the promising performance of our model on the
small datasets demonstrate the representational ability our method
compared to word-level and character-level embedding methods. On the
other hand, the advantages of Sent2Matrix on large datasets are even
more remarkable than that on small datasets. This indicates that the
Sent2Matrix embedding method enables CNN to apply 2-D filters to
compute more advanced features, thereby leading to better
generalization.

\section{Conclusions}

In this work, we propose the Sent2Matrix embedding method for
text representation. Our proposed method can overcome the
limitations of word-level and character-level embedding methods.
Sent2Matrix embedding method encodes sentences into
two-dimensional representations, thereby enabling CNNs to capture
both word-word and character-character relationships
simultaneously. To cope with variable-length words in sentences,
we develop the serpentine padding strategy, which retains the
sequential flow of character stream in sentences. Experimental
results on text classification tasks demonstrate that our new
embedding method with serpentine padding consistently outperforms
prior embedding methods.

\ifCLASSOPTIONcompsoc
  \section*{Acknowledgments}
\else
  \section*{Acknowledgment}
\fi

This work was supported in part by National Science Foundation
grants IIS-1908198 and IIS-1908166.

\ifCLASSOPTIONcaptionsoff
  \newpage
\fi


\bibliographystyle{IEEEtran}
\bibliography{deep}

\end{document}